\documentclass[sigconf, screen, svgnames]{acmart}

\usepackage{amsmath}
\usepackage{adjustbox}

\usepackage{tabularx}
\usepackage{subfig}

\usepackage{siunitx}
\usepackage{tikz}
\usetikzlibrary{arrows.meta,
                calc,
                decorations.pathreplacing,
                decorations.text,
                bending,
                fit,
                intersections,
                matrix,
                positioning,
                shapes,
                backgrounds}

\usepackage{array}
\newcolumntype{?}[1]{!{\vrule width #1}}

\usepackage{balance}

\usepackage[subtle]{savetrees}

\AtBeginDocument{%
  \providecommand\BibTeX{{%
    \normalfont B\kern-0.5em{\scshape i\kern-0.25em b}\kern-0.8em\TeX}}}

\copyrightyear{2021}
\acmYear{2021}
\setcopyright{acmcopyright}\acmConference[MM '21]{Proceedings of the 29th ACM International Conference on Multimedia}{October 20--24, 2021}{Virtual Event, China}
\acmBooktitle{Proceedings of the 29th ACM International Conference on Multimedia (MM '21), October 20--24, 2021, Virtual Event, China}
\acmPrice{15.00}
\acmDOI{10.1145/3474085.3475586}
\acmISBN{978-1-4503-8651-7/21/10}

\begin{document}

\title[Graph Neural Networks for Knowledge Enhanced Visual Representation of Paintings]{Graph Neural Networks for Knowledge Enhanced Visual\\ Representation of Paintings}

\author{Athanasios Efthymiou, Stevan Rudinac, Monika Kackovic, Marcel Worring, Nachoem Wijnberg}
\affiliation{%
  \institution{University of Amsterdam}
  \city{Amsterdam}
  \country{The Netherlands}
}
\email{{a.efthymiou, s.rudinac, m.kackovic, m.worring, n.m.wijnberg}@uva.nl}

\renewcommand{\shortauthors}{Efthymiou, et al.}

\begin{abstract}
We propose ArtSAGENet, a novel multimodal architecture that integrates Graph Neural Networks (GNNs) and Convolutional Neural Networks (CNNs), to jointly learn visual and semantic-based artistic representations. First, we illustrate the significant advantages of multi-task learning for fine art analysis and argue that it is conceptually a much more appropriate setting in the fine art domain than the single-task alternatives. We further demonstrate that several GNN architectures can outperform strong CNN baselines in a range of fine art analysis tasks, such as style classification, artist attribution, creation period estimation, and tag prediction, while training them requires an order of magnitude less computational time and only a small amount of labeled data. Finally, through extensive experimentation we show that our proposed ArtSAGENet captures and encodes valuable relational dependencies between the artists and the artworks, surpassing the performance of traditional methods that rely solely on the analysis of visual content. Our findings underline a great potential of integrating visual content and semantics for fine art analysis and curation.
\end{abstract}



\maketitle

\section{Introduction}
\label{introduction}

Fine art analysis has been a subject of intensive research in recent years. Advances in multimedia and related disciplines and the vast amount of publicly available artistic data have encouraged research in many fine art analysis tasks, ranging from artistic style classification and creation period estimation~\cite{elgammal_art_history_machine, tan_artgan, menisnk_rijksmuseum, strezoski_omniart}, style transfer~\cite{gatys_nst, jing_nst_review}, object detection and retrieval in paintings~\cite{crowley_object_retrieval_paintings, weakly_supervised_object_detection_artworks} to identification of semantic relationships between the artworks~\cite{celtinic_art_history_cnn, celtinic_dl_art}. However, most extant research concentrates almost exclusively on visual content analysis or in some cases on semantic associations.  In this study, we focus on capturing and modeling the complex visual and semantic relationships between artists and artworks to gain a deeper and more comprehensive understanding of paintings.

\begin{figure}[t]
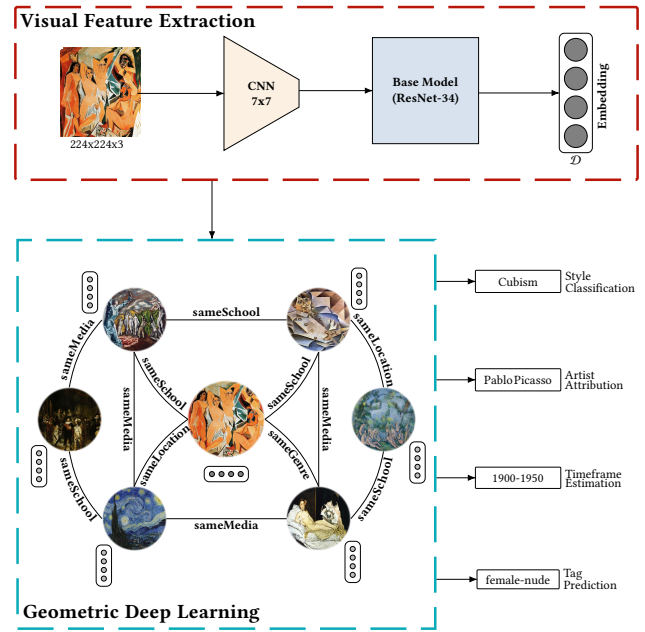

\centering
\input Figures/visual2semantics_grey_embeddings.tex
\caption{An illustration of our graph structured fine art analysis pipeline. First we extract visual features for node representations. Then, we utilize graph structure to learn context-aware fine art representations. Each node denotes a painting. Edges are drawn based on painting properties.
}
\label{fig:undirected_fine_art_knowledge_graph}
\end{figure}

Even though the visual content is a predominant characteristic of a visual artwork, automatic fine art analysis can significantly benefit from the well-established fundamental theories in the art domain, as well as the rich semantic information associated with the artworks. Personal links between painters, joint membership of artistic schools, and shared exhibitions are possible ties in such networks. By connecting visual content and semantic information representing such network relations, it is possible to better understand the processes of artistic innovation and influence. For instance, our understanding of Pablo Picasso’s proto-cubist seminal painting \textit{Les Demoiselles d’Avignon} (1907), depicted in Figure~\ref{fig:undirected_fine_art_knowledge_graph}, is greatly enhanced by recognizing the influence of Doménikos Theotokópoulos' (better known as El Greco) \textit{Opening of the Fifth Seal} (1608–1614) (upper left node in Figure~\ref{fig:undirected_fine_art_knowledge_graph}), or Édouard Manet’s \textit{Olympia} (1863) (lower right node)~\cite{the_girls_from_avignon_influences}.

Recent developments in structural modeling and the emergence of Graph Neural Networks (GNNs)~\cite{kipf_graph_cnn, graphsage, graphsaint} enable us to model these interesting properties and relationships. Nevertheless, most of the earlier GNNs fail to scale to domains that consist of large graphs with thousands of nodes and edges. In this work, we employ recently developed efficient GNN methods that can scale to large graphs with thousands of nodes and millions of edges~\cite{graphsage, graphsaint, chiang2019cluster, sign} and show that they are superior to traditional Convolutional Neural Networks (CNNs) in terms of predictive performance and computational efficiency for fine art analysis. To that end, we propose ArtSAGENet, a novel approach that extends GNNs and CNNs to jointly learn visual and semantic representations of fine art using the visual content of an artwork accompanied with its respective semantic relationships.

Following the rapidly growing body of research~\cite{ubernet, multi_task_survey, which_tasks}, we, first, employ a Multi-task Learning (MTL)~\cite{caruana_multitask} approach to learn visual representations of fine art. We argue that, next to the significant advantages of MTL in terms of computational resources, machines can substantially benefit from learning task inter-dependent representations of paintings. For example, even a non-expert (human) viewer of Pablo Picasso's \textit{Les Demoiselles d’Avignon} can easily recognize that the painting belongs to the Cubism stylistic movement, hence, it is more likely to be attributed to Pablo Picasso and created during the first half of the twentieth century. In addition to the multi-task visual representation learning, we introduce a knowledge-enhanced component that utilizes a GNN to model graph structured relationships among artists. Finally, we evaluate the performance of our proposed ArtSAGENet using the WikiArt dataset \cite{www.wikiart.org}, which has been extensively used for fine art classification tasks~\cite{elgammal_art_history_machine, celtinic_art_history_cnn, tan_artgan}. Experimental results show that ArtSAGENet outperforms several strong baselines and obtains state-of-the-art performance in fine art analysis, while qualitative analysis of the representations learned by our approach indicates that it is capable of capturing interesting properties of fine art. 

The main contributions of this work are the following:
\begin{itemize}
    \item We systematically evaluate multi-task learning for fine art analysis and confirm its significant advantages over single-task alternatives in a wide range of settings.
    \item We employ several Graph Neural Networks directly for a wide range of fine art analysis tasks by coupling visual features and homogeneous topological structure.
    \item We propose ArtSAGENet, a novel architecture for integrating semantic information and visual content.
    \item Through extensive experimentation we show that our proposed ArtSAGENet consistently outperforms strong traditional methods on style classification, artist attribution, creation period estimation and tag prediction\footnote{All models implementation and the dataset used are publicly available in \href{https://github.com/thefth/ArtSAGENet}{https://github.com/thefth/ArtSAGENet}.}.
\end{itemize}

The rest of this paper is structured as follows. In the next section, we review the relevant literature in automated fine art analysis, and thereafter, in Sections~\ref{approach} and~\ref{experimental_setup} we provide all the necessary details for our proposed methods and the experimental setup. Finally, in Section~\ref{results} we present the experimental results, and in Section~\ref{conclusion} we conclude discussing our findings.  

\section{Related Work}
\label{related_work}

\begin{figure*}[t]
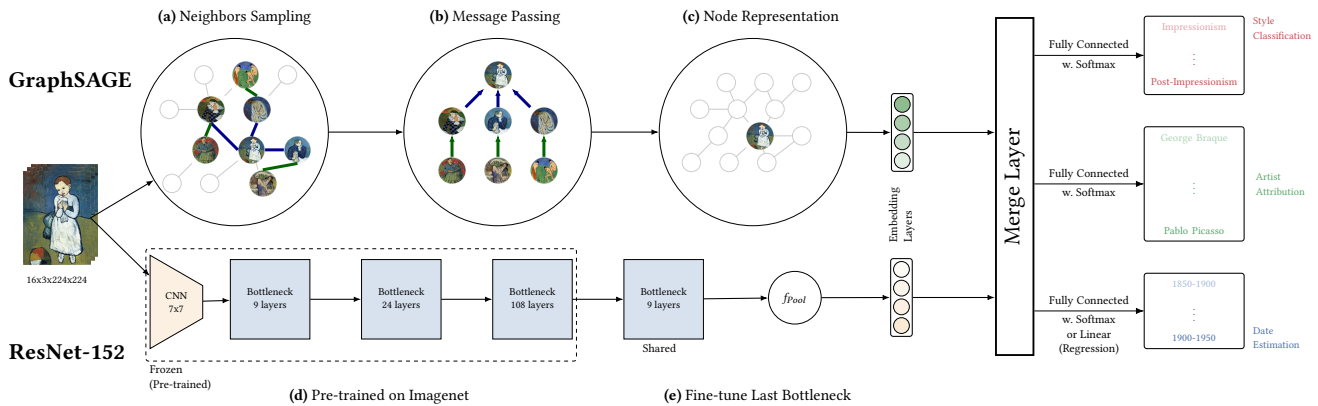

\centering
\input Figures/artsagenet_architecture.tex
\caption{This figure depicts the  ArtSAGENet architecture. Given a batch of images, the forward propagation is implemented as follows. GraphSAGE: (a) For each image in the batch we sample \textit{k} neighbors, e.g., $k=3$, from \textit{h} hops, e.g., $h= 2$, (b) aggregate the node feature vectors within the neighborhood to (c) obtain the final node representation. ResNet: (a) Each image of the batch is passing through the frozen part of the pre-trained (on ImageNet) network and then (b) it passes through the network's last bottleneck which is trainable and fine-tuned to obtain the final visual representation. Finally, the obtained representations are merged and the final multimodal representation is passed to the last layer (classifier or regressor). GraphSAGE (a-c) and ResNet-152 (e) are jointly trained for all three tasks in an MTL manner.
}
\label{fig:resnet_152_graphsage}
\end{figure*}

The recent successes in deep learning for multimedia have inspired research in fine art classification based on visual \cite{cetinic_understanding, deepart, elgammal_art_history_machine, menisnk_rijksmuseum, strezoski_multitask} and semantic content \cite{context_aware_embeddings, celtinic_art_history_cnn, kim_semantics_art, semantic_art_understanding}. In this section we briefly survey both research lines as well as multimodal approaches that integrate visual content and semantics.

\subsection{Visual Content-Based Fine Art Analysis} \label{subsec:visual_content_rw}
Visual content-based fine art analysis has attracted a lot of attention following the recent advancements in computer vision and the emergence of large-scale visual art collections that contain paintings, sculptures, photographs and installations, such as the Rijksmuseum~\cite{menisnk_rijksmuseum}, DeepArt \cite{deepart}, BAM! \cite{wilber_bam} and OmniArt \cite{strezoski_omniart} datasets. \citet{strezoski_artsight} develop an interactive method to explore visual art collections using colors as visual cues, while the web-based platform proposed in~\cite{strezoski_ace} facilitates interactive exploration of the visual sentiment and emotion in paintings over time. ~\citet{shen_visual_patterns_art_collections} propose self-supervised spatially-consistent feature learning to discover near duplicate patterns in large collections of artworks. \citet{recognizing_style} learn latent style representations inspired by the Gram matrix based correlation calculation using the VGG architecture ~\cite{vgg} and evaluate their proposed methods using the Painting91~\cite{khan_painting_91}, arcDataset~\cite{xu_arcdataset} and Hipster Wars~\cite{kiapour_hipster_wars} datasets.

\citet{bianco_multitask} propose a Spatial Transformer Network to identify the most discriminative sub-regions of paintings by employing CNNs and evaluate the performance of their approach by constructing the MultitaskPainting100k dataset. Similarly, \citet{celtinic_dl_art} utilize CNNs to study aesthetics, memorability and sentiment in paintings, while \citet{celtinic_art_history_cnn} and \citet{elgammal_art_history_machine} explore the WikiArt collection to quantify stylistic properties according to Heinrich W{\"o}fflin's concepts  \cite{wofflin_principles_art_history}. \citet{strezoski_multitask} employ a multi-task setting for several fine art classification tasks ranging from artist attribution and type prediction to period estimation. \citet{tan_artgan} utilize AlexNet and evaluate its performance on the artist attribution task amongst others. In contrast to these works, we don't rely solely on traditional CNN architectures, but we model complex semantic-based artistic relationships that go beyond visual content to facilitate context-aware art analysis.

\subsection{Semantic Content-Based Fine Art Analysis} \label{subsec:semantic_content_rw}

There is an emerging interest in studying semantic principles of the fine art history~\cite{kim_semantics_art, celtinic_art_history_cnn}. \citet{principals_semantics} utilize CNNs trained for style classification and study the intermediate layer's activations for semantic interpretation. \citet{semantic_art_understanding} develop SemArt and propose the Text2art challenge to evaluate semantic art understanding. In their work, they collect artistic data from the Web Gallery of Art (WGA) containing a textual description of each painting as well as  information about the author.

\citet{artpedia} address the problem of cross-modal retrieval of paintings and their associated descriptions, and create the Artpedia dataset that consists of nearly three thousands paintings annotated with contextual sentences. \citet{recognizing_characters} employ traditional machine learning methods, and show that CNNs can efficiently learn art-specific domain knowledge to recognize human figures in paintings. \citet{artgcn} propose ArtGCN to learn artistic node representations based on paintings' textual descriptions. They compare their ArtGCN with visual content CNN baselines~\cite{resnet} using SemArt~\cite{semantic_art_understanding}, and show that their proposed method can surpass CNNs performance. In contrast, we do not employ explicit textual information, but we focus on the significant advantages of integrating visual content and semantic-based information.

\subsection{Multimodal Representations} \label{subsec:mml_rw}

Recently, multimodal modeling for integrating visual content and semantics has gained a lot of traction. Our work is related to that of \citet{mlgcn}. They propose multi-label Graph Convolutional Neural Networks by utilizing a multimodal architecture that jointly learns image representations and object label inter-dependencies using Graph Convolutional Networks~\cite{kipf_graph_cnn}, achieving state-of-the-art performance in object recognition tasks. In contrast to that, here we employ recently developed efficient Graph Neural Networks~\cite{chiang2019cluster, graphsage, graphsaint, sign} that can scale to large graphs with thousands of nodes and learn from the semantic relationships between paintings for fine art categorization tasks.

\citet{context_aware_embeddings} adopt the node2vec~\cite{node2vec} framework and build a directed heterogeneous graph using several artistic properties, such as artist schools, artwork materials and titles to jointly learn visual and contextual representations of fine art.  They evaluate their proposed method using SemArt~\cite{semantic_art_understanding}. In contrast to our work, \citet{context_aware_embeddings} introduce intermediate nodes of other types to encode semantic relations between the paintings and do not directly use the node2vec framework for fine art node classification, but rather employ the graph structure to learn contextualized visual embeddings as in the neural structure learning paradigm \cite{neural_structure_learning}.
Finally, \citet{hyperlearn} propose HyperLearn, a method that learns hypergraph structures and employ their approach on artist attribution task achieving comparable results to OmniArt benchmark~\cite{strezoski_omniart}, while using only a small portion of labeled data. Inspired by this line of research, we explore the potential of recently proposed GNNs originally designed for homogeneous graph structures. However, contrary to related work on employing GNNs for fine art analysis, e.g., \cite{hyperlearn, context_aware_embeddings}, we utilize an undirected homogeneous graph structure where we consider only paintings as nodes connected based on a single shared attribute.

\section{Learning knowledge-enhanced visual representations}
\label{approach}

Traditional CNNs have been proven extremely effective in fine art analysis. However, the visual arts domain is characterized by an exceptional information richness, including e.g., the explicit and implicit social networks between artists. To that end, in this work we take a multimodal approach to enhance the performance of the conventional visual content-based CNNs. Figure \ref{fig:resnet_152_graphsage} depicts ArtSAGENet a two-branch architecture that jointly learns a CNN and a GNN-based model to learn multimodal deep context-aware visual representations of paintings.

\subsection{Visual Representation Learning}

Even though an arbitrary CNN can be deployed for learning visual representations, in this work we are adopting the ResNet-152~\cite{resnet} architecture due to its excellent performance, choice which we will be further motivated in Section~\ref{subsec:baselines}. Adopting the same notation as in \cite{mlgcn}, for a given painting, we can obtain a visual representation $v_i$ as follows:
\begin{align} \label{al:visual_embedding}
    v_i &= F_{GAP}(f_{CNN}(\mathcal{P};\theta_{CNN})) \in \mathbb{R}^D,
\end{align}
where  $\mathcal{P}$ denotes the painting's photographic reproduction, $F_{GAP}$ denotes the global average pooling operation, $\theta_{CNN}$ denotes the ResNet-152, or any other CNN-based model parameters, and $D$ denotes the visual embedding dimensionality, i.e., 2,048. We fine-tune the last bottleneck of a pre-trained on ImageNet \cite{imagenet} ResNet-152, albeit we omit the classifier. For either classification or regression tasks, we merge the learned visual and contextual embeddings to obtain the final multimodal painting representation.

\subsection{Graph Representation Learning} \label{subsec:approach_gnns}

Recently, Graph Neural Networks have gained popularity, because of their outstanding performance in node classification, graph classification and link prediction tasks. Given the nature of the visual arts domain, semantic knowledge can be of great leverage when analyzing artists and artworks. We conjecture that information about, e.g., social networks between the artists, if encoded properly, can be exploited by GNNs for improved fine art analysis.

To this end, we propose a novel method of classifying artwork attributes following the node classification paradigm, as shown in Figure \ref{fig:undirected_fine_art_knowledge_graph}. That is, we treat each artwork as a node with links to other artworks/nodes based on their semantic properties.

Early GNN methods usually operate on an adjacency matrix $\mathcal{A}$ that encodes these relations between nodes in a full-batch setting. This significantly increases the computation complexity that hinders scaling to large graph structures. Recent GNNs propose a number of sampling methods that alleviate the latter issue. Here, we employ such GNN architectures~\cite{chiang2019cluster, graphsage, graphsaint, sign} that can scale to homogeneous graph structures with thousands of nodes and millions of edges.

\textbf{Graph construction.}
Before training the aforementioned GNNs, we need to construct a predefined adjacency matrix $\mathcal{A}$. We build the adjacency matrix $\mathcal{A}$ using the following:
\begin{align} \label{al:graph}
\mathcal{A}_{ij} &=
\begin{cases}
 1,         & \footnotesize \text{if $property(artwork_i)=property(artwork_j)$}  \\
 0,         & \footnotesize \text{if $property(artwork_i) \neq property(artwork_j)$}
\end{cases}
\end{align}
Since, most of the GNN methods that we adopt were originally developed for learning in undirected homogeneous graph structures, we build an undirected homogeneous graph accordingly. We leave working with recently proposed GNNs~\cite{heterogeneous_gnn, heterogeneous_gat} that can leverage structural heterogeneity for future research. We follow \cite{hyperlearn, context_aware_embeddings} and use artistic schools to link the artworks nodes, e.g., all of the available Paul C\'ezanne's  paintings will be linked with the ones of \'Edouard Manet's given that both artists represent the French school of painting. We follow the same practice as in~\cite{graphsage} and downsample the edges in the original graph, so that nodes can have at most a degree of 128.

For our ArtSAGENet architecture we utilized the GraphSAGE architecture~\cite{graphsage} to obtain context-aware node representations. Figure \ref{fig:resnet_152_graphsage} illustrates the proposed method. That is, given a node we uniformly sample $k$ neighbors from $h$ hops and aggregate the neighborhood's node feature vectors to obtain the node representation. We use a mean aggregator and obtain the node representation $n_i$ as follows:
\begin{align} \label{al:node_representation}
    {n_i} &= W_1 x_i + W_2 \cdot f_{AGG} x_j, \forall j\in \mathcal{N}(i),
\end{align}
where $x_i$ is the node feature vector, $\mathcal{N}(i)$ denotes the neighborhood for the node $i$, $x_j$ the node feature vector of the neighbor $j$, $f_{AGG}$ the neighbors feature vectors aggregator, while $W_1$ and $W_2$ the learned weights. We followed~\cite{graphsage} and set the number of hops $h=2$ with neighborhood sample sizes $k_1=25$ and $k_2=10$ for the hops $h_1$ and $h_2$, respectively.

We considered two methods to obtain the node feature vectors. 

\textbf{Visual features.} First, we utilize visual representations as node features for the proposed GNN model in a multimodal fusion manner. That is, prior to training the GNNs, we utilize a pre-trained on ImageNet and frozen ResNet-34 architecture as a backbone to extract 512-dimensional paintings visual feature vectors. In particular, we extract the features after the last convolutional layer. Thereafter, we train the proposed GNNs using the image-level visual feature vector $v_i$, as computed in Eq. \eqref{al:visual_embedding}. 

\textbf{Bag-of-words tag feature vectors.} We also consider sparse input features. To this end, we use the painting's tags as node feature vectors in a bag-of-words manner, i.e., by representing each node as an one-hot encoded vector based on its attributed tags. We collected the tags associated with the artworks from the WikiArt online collection. We considered only the 1,170 tags that appear more than 10 times in the WikiArt collection and introduce a special tag \textit{Unknown} for paintings with no available tags.

\subsection{Multimodal Embedding}

Given the learned visual and context-aware embeddings, we use a merge operation to get the final knowledge enhanced visual representation. Even though we considered different merging operations, in this work we report results using only the concatenation of the visual and semantic embeddings as follows:
\begin{align} \label{al:mml_embedding}
    x_i = v_i \oplus n_i
\end{align}

Comparative results using alternate algebraic merge operations can be found in supplementary materials.

\subsection{Multi-Task Learning}

Multi-task learning (MTL) is the setting of training an algorithm over multiple tasks and has been shown to be extremely suitable for fine art analysis \cite{context_aware_embeddings, strezoski_multitask}. In this work, we propose and evaluate an MTL setting in order to attribute artworks to stylistic movements, artists and creation periods. We argue, that these three specific tasks are highly cooperative tasks, and therefore MTL can enhance the performance of our proposed methods. We use the following formula for training:
\begin{align} \label{al:mtl}
    L_T &= \sum_{t=1}^T w_t \mathcal{L}_t,
\end{align}
where $w_t$ denotes the task-specific weight, $\mathcal{L}_t$ the loss for task $t$ and $L_T$ the combined loss across all tasks $T$. In Section~\ref{results}, we report results using the same task-specific weight $w_i$ for each task $t$.

For multi-class classification, we employ the categorical cross-entropy loss:
\begin{align} \label{al:cross_entropy}
    \mathcal{L} &= - \frac{1}{N} \sum_{k=1}^N \sum_{c=1}^C y_k \log \hat{y}_k,
\end{align}
where $N$ denotes the total number of the samples considered, $C$ the number of total classes, $y_k$ the ground-truth label, and $\hat{y}_k$ the output given by the \textit{softmax} function for the sample $k$.

\section{Experimental Setup}
\label{experimental_setup}

In this work, we utilize traditional visual content CNNs~\cite{alexnet, vgg, resnet} and recently proposed GNNs~\cite{chiang2019cluster, graphsaint, graphsage, sign} as our baselines. In this section, we present the data collection and tasks that we use, and provide all necessary implementation details.

\subsection{Dataset Variants and Tasks} \label{subsec:dataset}

We evaluate our proposed methods using the WikiArt online user-editable visual arts collection on four downstream tasks, namely style classification, artist attribution, creation period estimation and tag prediction. 
For a fair comparison with related works~\cite{elgammal_art_history_machine, celtinic_fine_tuning_cnn, tan_artgan} that used the WikiArt collection, we evaluate our proposed methods using the WikiArt\textsuperscript{Full} dataset that consists of 75,921 paintings, and the WikiArt\textsuperscript{Artists} subset that considers only the works of the 23 most representative artists in the WikiArt collection. In addition, we introduce another subset WikiArt\textsuperscript{Modern}, that consists of all the artworks from the year 1850 up to 1999, since we are interested in the modern art period as it incorporates the emergence of many major stylistic movements and several groundbreaking moments in the history of fine art. We make use of the WikiArt\textsuperscript{Modern} subset to evaluate the performance of our proposed methods in creation year estimation task. For this regression problem, we employ the Mean Absolute Error (MAE)~\cite{age_estimation_mae} for training. Inspired by research in human age estimation~\cite{age_estimation_cumulative_score, age_estimation_bridge_net,age_estimation_mean_variance}, we report the Cumulative Score (CS) \cite{age_estimation_cumulative_score} evaluation metric, which is defined as follows:
\begin{align} \label{al:cumulative_score}
    CS(\theta) &= \frac{N_\theta}{N}\times100,
\end{align}
where $N$ is the total number of paintings in the test set and $N_\theta$ denotes the number of paintings whose absolute error is less than $\theta$ years. Table~\ref{tab:dataset_statistics} summarizes the dataset statistics for all the WikiArt dataset variants that we use in this work. Note that in Table~\ref{tab:dataset_statistics} timeframes designate half-century periods, such as 1900-1950.

Finally, we evaluate the performance of all methods in a multi-label tag prediction task. To this end, we make use of 54,919 paintings that are associate with at least one tag that, in turn, appears at least in 1,000 unique paintings. This results in 54 unique tags, ranging from face parts, e.g., forehead and lips, nature related tags, e.g., flowers and animals, urban objects, e.g., vehicles and boats, to themes, e.g., children portraits and famous people. A detailed description of the dataset collection is provided in supplementary materials.

\begin{table}[t]
    \centering
    \caption{Statistics of the WikiArt collection variants.}
    \label{tab:dataset_statistics}
    \begin{tabular*}{\columnwidth}{l @{\extracolsep{\fill}} S[table-parse-only,table-text-alignment=left,
                        table-number-alignment=left]S[table-parse-only, table-text-alignment=left,
                        table-number-alignment=left]S[table-parse-only, table-text-alignment=left,
                        table-number-alignment=left]}
    \toprule
       Attribute  & WikiArt\textsuperscript{Full} & WikiArt\textsuperscript{Modern} & WikiArt\textsuperscript{Artists} \\\midrule
        Artworks & 75921 & 45869   & 17785   \\
        Artists  & 750 & 462  & 23 \\
        Styles  & 20 & 13  & 12 \\
        Dates  & 587 & 150 & 240  \\
        Timeframes  & 13 & 4 & 8 \\
        Tags & 4879 & 3652  & 2370 \\\bottomrule
    \end{tabular*}
\end{table}

\subsection{Baselines} \label{subsec:experimental_setup_baselines}

We follow the approach in~\cite{elgammal_art_history_machine} and evaluate AlexNet~\cite{alexnet}, VGG~\cite{vgg} and ResNet~\cite{resnet} architectures and their variants as CNN baselines. For model implementation, training and evaluation we used the PyTorch library~\cite{pytorch}. For all CNN architectures, we are fine-tuning the last convolutional layer alongside the final fully-connected layer(s) by using pretrained versions on ImageNet.

We employ four recently proposed Graph Neural Networks that perform remarkably well in node classification tasks, namely Cluster-GCN~\cite{chiang2019cluster}, GraphSAGE~\cite{graphsage}, GraphSAINT~\cite{graphsaint} and SIGN~\cite{sign}, as GNN baselines. We use the same graph as constructed in Eq. \eqref{al:graph}. In addition, for all GNN baselines we use the paintings visual feature vectors as node representations. We extract these visual feature vectors from the last convolutional layer of a ResNet-34 pre-trained on ImageNet as in Section~\ref{subsec:approach_gnns}. An illustration of the aforementioned approach is shown in Figure~\ref{fig:undirected_fine_art_knowledge_graph}. For GNN architectures implementation we used the PyTorch Geometric library \cite{pytorch_geometric}. We experimented with several configurations for each GNN and always selected the best performing variant.

\subsection{Implementation Details} \label{subsec:implementation}

For consistency with previous work, we using the same dataset split scheme as in~\cite{elgammal_art_history_machine}. That is, we use the 85\% of the dataset as training set, the 9.5\% as validation set and the remaining 5.5\% as test set. For the GNN baselines and our ArtSAGENet we consider node neighbors only from the same set, and we omit edges between paintings from the same artist in the first hop. When training any CNN architecture, we augment the input images by using a random horizontal flip with a given probability $p=0.5$.

For all models, we use early stopping by monitoring the validation error, i.e., we stop training if there is no improvement in validation loss after ten consecutive epochs. In addition to early stopping, we adopt a dynamic learning rate reducing strategy. That is, we monitor the validation error and reduce the learning rate by a factor of 10 once learning stagnates, i.e., if there is no improvement in validation loss after five consecutive epochs. We experimented with various settings for all models hyperparameters. We train all CNN architectures and our ArtSAGENet using Stochastic Gradient Descent (SGD)~\cite{sgd} with initial learning rate set to 0.001 and momentum to 0.9 with a mini-batch size of 16. For GNNs, we use the Adam optimizer~\cite{adam} with a learning rate of 0.001 and a mini-batch size of 1,024, but we omit the dynamic learning rate scheduler.

\section{Results}
\label{results}

In this section, we present the experimental and qualitative evaluation of our proposed method for fine art analysis.

\begin{table*}[t]
\centering
\caption{Accuracy of ArtSAGENet and the baselines. MTL denotes Multitask Learning. WikiArt\textsuperscript{Modern} - Date$\ddagger$ reports the cumulative score as in Eq. \eqref{al:cumulative_score} with $\theta=5$ years. $\clubsuit$ means using tags as node feature vectors. $\spadesuit$ means using features extracted from a ResNet-34 model pre-trained on ImageNet as node feature vectors. Higher is better (best results in \textbf{bold}).}
\label{tab:results}
\begin{tabular*}{\linewidth}{l @{\extracolsep{\fill}} ccccccccc}
\toprule
                & \multicolumn{3}{c}{WikiArt\textsuperscript{Full}}                                                     & \multicolumn{3}{c}{WikiArt\textsuperscript{Modern}}                                                       & \multicolumn{3}{c}{WikiArt\textsuperscript{Artists}}                                                          \\ \cmidrule( r){2-4}
\cmidrule(lr){5-7}
\cmidrule(l ){8-10}
Model                    & Style & Artist & Timeframe & Style & Artist & Date$\ddagger$ & Style & Artist & Timeframe \\ \midrule
AlexNet     &     60.3          &     44.0       &   65.8           &      62.6       &      46.1        &   23.2         &      77.8         &    76.9       &       79.1                     \\
AlexNet - MTL   &     61.8 & 45.7 & 66.6        &    63.3 & 47.3 & 18.8           &  79.5 & 78.6 & 79.9                          \\
VGG-16      &      68.3         &    58.5        &   71.1             &      68.9       &     61.2          &      23.0      &       83.8        &   85.6         &    81.2                    \\
VGG-16 - MTL     &             70.0 & 57.2 & 73.3                &      69.7 & 55.7 & 22.9         &             82.8 & 86.2 &  82.4                          \\
VGG-19     &       67.7                  &    55.2            &    70.2         &  67.6            &    57.8        & 23.9 &   82.1          &   85.4         &     79.6                   \\
VGG-19 - MTL     &         67.9 & 54.9 & 71.5          &           68.3 & 52.6 & 22.8    &   84.2 & 84.4 & 84.2                   \\
ResNet-34 &    67.2           &   59.6         &    70.0            &    68.0         &     61.9         &  23.5           &    79.7           &     86.3       &         78.7                  \\
ResNet-34 - MTL  &            69.5 & 58.1 & 73.2              &         71.1 & 58.3 & 22.9  &         83.5 & 85.9 & 83.4                          \\
ResNet-152 &    69.3           &   64.1        &    72.3            &    69.9        &      66.1        & 23.2           &   84.4            &   88.6         &   83.2                       \\
ResNet-152 - MTL & 74.0 & 62.3 & 76.5              &             73.1 & 64.6 & 25.1       & 85.3 &  88.6 & 86.1                          \\\midrule
Cluster-GCN           &        69.6            &   57.9             &     71.3        &    67.8          &   58.8         & 21.7              &    86.4        &        94.5        &  85.2           \\
GraphSAGE           &            69.9          &     60.8           &    73.0        &    69.5          &  65.7         &  23.8          &    \textbf{88.5}           &   \textbf{99.0}            &  85.7           \\ 
GraphSAINT           &         71.2            &    62.4           &      72.9     &      70.0        &  65.3          &   \textbf{31.7}            &  88.4          &         98.1       &   86.4          \\
SIGN           &           70.1             &       65.5        &    71.4         &   69.1            &   67.1          &     22.4           & 81.6            &  96.9             &    83.5        \\ \midrule
ArtSAGENet\textsuperscript{$\clubsuit$}  - MTL           &  76.1 & 73.9 & 77.7                    &  74.5 & 66.9 & 24.3       &  85.2 & 93.4 & 87.3                \\
ArtSAGENet\textsuperscript{$\spadesuit$}  - MTL                 &  \textbf{77.6} & \textbf{76.6} & \textbf{79.2}                    &   \textbf{76.7} & \textbf{75.4} & 24.2       &  \textbf{88.5} & 98.1 & \textbf{88.4}                 \\\bottomrule
\end{tabular*}
\end{table*}

\begin{figure*}[t]
    \centering
    \includegraphics[width=\textwidth]{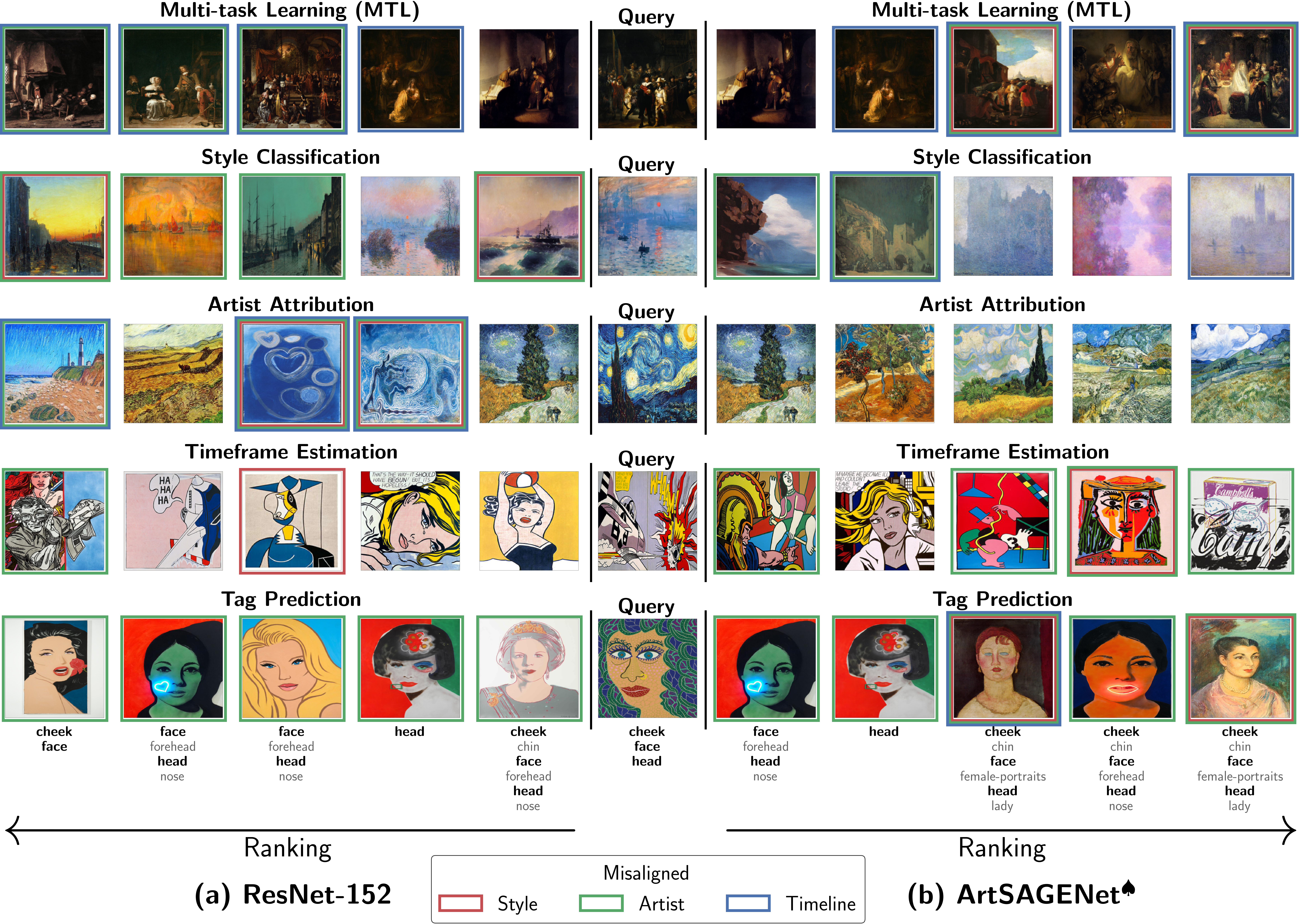}
    \caption{Qualitative analysis of learned visual representations for painting retrieval. Given a reference painting (middle), the top-5 nearest neighbors of the ResNet-152 (left) and the ArtSAGENet were retrieved (right). Misaligned patches denote paintings attributed with different \textcolor[HTML]{C44E52}{style}, \textcolor[HTML]{55A868}{artist} or \textcolor[HTML]{4C72B0}{timeline} annotation(s) from the reference painting. MTL illustrates the top-5 nearest neighbors retrieved using the Multi-task Learning model trained for style classification, artist attribution and timeframe estimation. The rest of the rows illustrate the top-5 nearest neighbors retrieved using the single-task classifiers. For the single-task tag prediction classifier, \textbf{bold} means that tag is attributed to the query painting, too. $\spadesuit$ means using visual features as node feature vectors.}
    \label{fig:qualitative_analysis_embeddings}
\end{figure*}

\subsection{Fine Art Categorization} \label{subsec:baselines}

\hspace{\parindent}\textbf{ResNet is the dominant CNN architecture}. Traditional CNN architectures have been proven to be extremely powerful for visual arts analysis obtaining state-of-the-art performance in style classification~\cite{celtinic_art_history_cnn, elgammal_art_history_machine} and artist attribution tasks~\cite{compare_the_performance_art_classification}. We observe a similar effect as in~\cite{elgammal_art_history_machine}. That is, for almost every task and dataset variant the dominant CNN is ResNet-152, followed by ResNet-34, while AlexNet seems to be the worst performer. Table~\ref{tab:results} summarizes the evaluation results for all models and dataset variants. 

Additionally, the dynamic learning scheduler that we have adopted seems to be extremely helpful, increasing the performance of our CNN baselines in all tasks. For example, single-task ResNet-152 achieves a 6\% boost in performance compared to previous work~\cite{elgammal_art_history_machine} on style classification on WikiArt\textsuperscript{Full}. Furthermore, the multi-task learning approach enhances the performance of all CNNs across almost all tasks. We note that ResNet-152 performance on WikiArt\textsuperscript{Full} is the state-of-the-art performance using visual content-based CNNs. Finally, we have experimented with training the models both from-scratch and fine-tuning them. We observe that fine-tuned models that are pre-trained on ImageNet consistently outperform their from-scratch counterparts, while they converge extremely faster. For that reason, in this paper we do not report results of training from scratch.

\textbf{Graph Neural Networks for fine art analysis.} Table \ref{tab:results} reports the performance of our GNN baselines. Consistent with the other studies, we observe that multi-task learning in some cases deteriorates GNNs performance, thus, we report results using single-task GNNs. Each GNN seems to perform on-par with the traditional single-task CNN models, while they clearly outperform AlexNet and VGG and obtain state-of-the-art performance on the smaller dataset variant WikiArt\textsuperscript{Artists}. 

We further note that GNNs achieve an outstanding performance on WikiArt\textsuperscript{Artists} artist attribution task. We hypothesize that this behavior is due to the simplicity of the task using the inherited artistic school attribute. Another interesting observation is that GraphSAGE requires only 20\% of the available labeled data for training to surpass the performance of ResNet-152 - MTL trained on the WikiArt\textsuperscript{Artists} for artist attribution task.

\textbf{ArtSAGENet obtains the state-of-the-art performance in fine art analysis.} Finally, we evaluate our proposed ArtSAGENet using either visual features or tags as node feature vectors. Since we observe a slight boost in performance, we report results using multi-task learning. Table~\ref{tab:results} illustrates that our proposed architecture is the dominant method for fine art analysis tasks. 

It can be clearly seen that both ArtSAGENet variants consistently outperform all baselines in style classification and timeframe estimation. In addition, ArtSAGENet outperforms all baselines on the WikiArt\textsuperscript{Full} and WikiArt\textsuperscript{Modern} artist attribution task, while it obtains the second best performance on the WikiArt\textsuperscript{Artists} dataset variant. We also have to note that using dense node visual feature vectors yields better performance than its sparse counterpart that relies on tags.

Another major observation is that ArtSAGENet significantly outperforms the multi-task ResNet-152 in the artist attribution task across all dataset variants. We hypothesize that the art school attribute is in general an informative property, which ArtSAGENet leverages to accurately attribute artists to artworks. Nevertheless, it seems to be inferior to GraphSAINT on WikiArt\textsuperscript{Modern} creation year estimation task within a $\pm10$ years period, albeit both methods perform comparably after a period of $\pm25$ years. The cumulative accuracy curves for the creation year estimation task are shown in supplementary materials.

\subsection{Multi-Label Tag Prediction} \label{subsec:tags_prediction}

In order to evaluate the performance of our proposed methods for multi-label fine art categorization, we employ a fine art tag prediction task. Inspired by work in object recognition~\cite{mlgcn, cnn_rnn}, we report per-class F1-score (CF1), overall F1-score (OF1) and mean Average Precision (mAP) for the tag categories. Detailed definitions of these measures are provided in~\cite{multi_label_performance}.

Results are summarized in Table~\ref{tab:tag_prediction_results}. Once again, ArtSAGENet is clearly the best performer. Yet another interesting observation is that strong CNNs, e.g., VGG and ResNet, perform better than their GNN counterparts. This suggests that a simple homogeneous structural topology that encodes information about the artistic school can not surpass the performance of powerful visual content-based methods in more complex tasks, as in tag prediction. However, it seems that all GNN baselines clearly outperform AlexNet in terms of mAP, while they need an order of magnitude less time for training and inference compared to CNN architectures.

In Table \ref{tab:time_comparison} we compare the time needed for training and inference in case of the CNN baselines, our ArtSAGENet and the GraphSAGE architecture. Since the computational runtime of GNNs is in the same order of magnitude, we highlight here only GraphSAGE, which we use as a building block for our ArtSAGENet. GraphSAGE clearly requires almost 50 times less time for training than any CNN. Finally, we observe that our ArtSAGENet requires only a small amount of time more than the ResNet-152 it relies on.

\begin{table}
\centering
\caption{Performance comparison on tag prediction task. $\spadesuit$ means using visual features as node feature vectors. Higher is better (best results in \textbf{bold}).}
\label{tab:tag_prediction_results}
\begin{tabular*}{\columnwidth}{l @{\extracolsep{\fill}}  ccc}
\toprule
Model      	& CF1 & 	OF1 & mAP \\ \midrule
AlexNet    	& 54.1 & 59.8 & 60.9         \\
VGG-16     	 & 57.4 & 63.6 & 65.3             \\
VGG-19     	& 56.8 & 62.9 & 64.9           \\
ResNet-34  	& 59.4 & 64.6 & 66.3  \\
ResNet-152 	 & 60.3	& 65.2 & 66.7              \\\midrule
Cluster-GCN 	& 49.6 	& 58.4 & 62.5     \\ 
GraphSAINT 	 & 55.4	 & 61.9 & 63.8      \\ 
GraphSAGE  	& 50.0 	& 58.9 & 63.3          \\ 
SIGN       	 & 51.7	& 59.6 & 63.7        \\ \midrule
ArtSAGENet\textsuperscript{$\spadesuit$}      &    \textbf{62.1} & \textbf{66.9} & \textbf{68.6}          \\ \bottomrule
\end{tabular*}
\end{table}

\begin{table}
\centering
\caption{
Time analysis relative to GraphSAGE (fastest) on tag prediction task. $\spadesuit$ means using visual features as node feature vectors. Train per epoch reports relative time difference for forward/backward operations for a single epoch for all models using the same batch size, i.e., 16. Inference reports the relative time difference for forward operations on test time for full test set. For each model and both training/inference we utilized an Nvidia GeForce 1080Ti 11GB GDDR5X GPU.}
\label{tab:time_comparison}
\begin{tabular*}{\columnwidth}{l @{\extracolsep{\fill}} ccr}
\toprule
         Model& Train (per epoch)& Inference & Parameters\\\midrule
         GraphSAGE&$\times$1 &$\times$1 &5M \\\midrule
         AlexNet&$\times$47.11 &$\times$2.86 & 57M\\
         VGG-16&$\times$53.14 &$\times$3.11 & 134M\\
         VGG-19&$\times$54.41 &$\times$3.19 &140M\\
         ResNet-34 & $\times$48.54 &$\times$2.89 &21M \\
         ResNet-152&$\times$52.60 &$\times$3.09 &58M\\\midrule
         ArtSAGENet\textsuperscript{$\spadesuit$}&$\times$52.74 &$\times$3.10& 63M \\\bottomrule
    \end{tabular*}
\end{table}

\subsection{Qualitative Analysis on Paintings Retrieval} \label{subsec:qualitative_analysis}

Further to fine art analysis tasks, we evaluate the performance of our ArtSAGENet in content-based fine art retrieval. That is, we extract the learned representations for both ResNet-152 and ArtSAGENet models and use the \textit{k-nearest neighbors} (k-NN) algorithm to retrieve the top-5 nearest neighbors for each case. For ResNet-152 we extract the learned visual representations as in Eq.~\eqref{al:visual_embedding}, while for our ArtSAGENet we use the multimodal embedding as obtained in Eq. \eqref{al:mml_embedding}. Then, for each reference painting we sort the paintings from the collection in ascending order based on their respective cosine distance to the query image.

Figure \ref{fig:qualitative_analysis_embeddings} illustrates the results for the ResNet-152 and the proposed ArtSAGENet models for each training setting, i.e., single-task versus multi-task learning. We can observe that most of the retrieved paintings have the same properties as the query painting for each method. However, we have to note that the single-task ArtSAGENet retrieval results for style classification, artist attribution and timeframe estimation are consistent across the top-5 nearest neighbors, i.e., all neighbors share the same attribute of interest with the query object.

\begin{figure}
\centering
\includegraphics[width=\columnwidth]{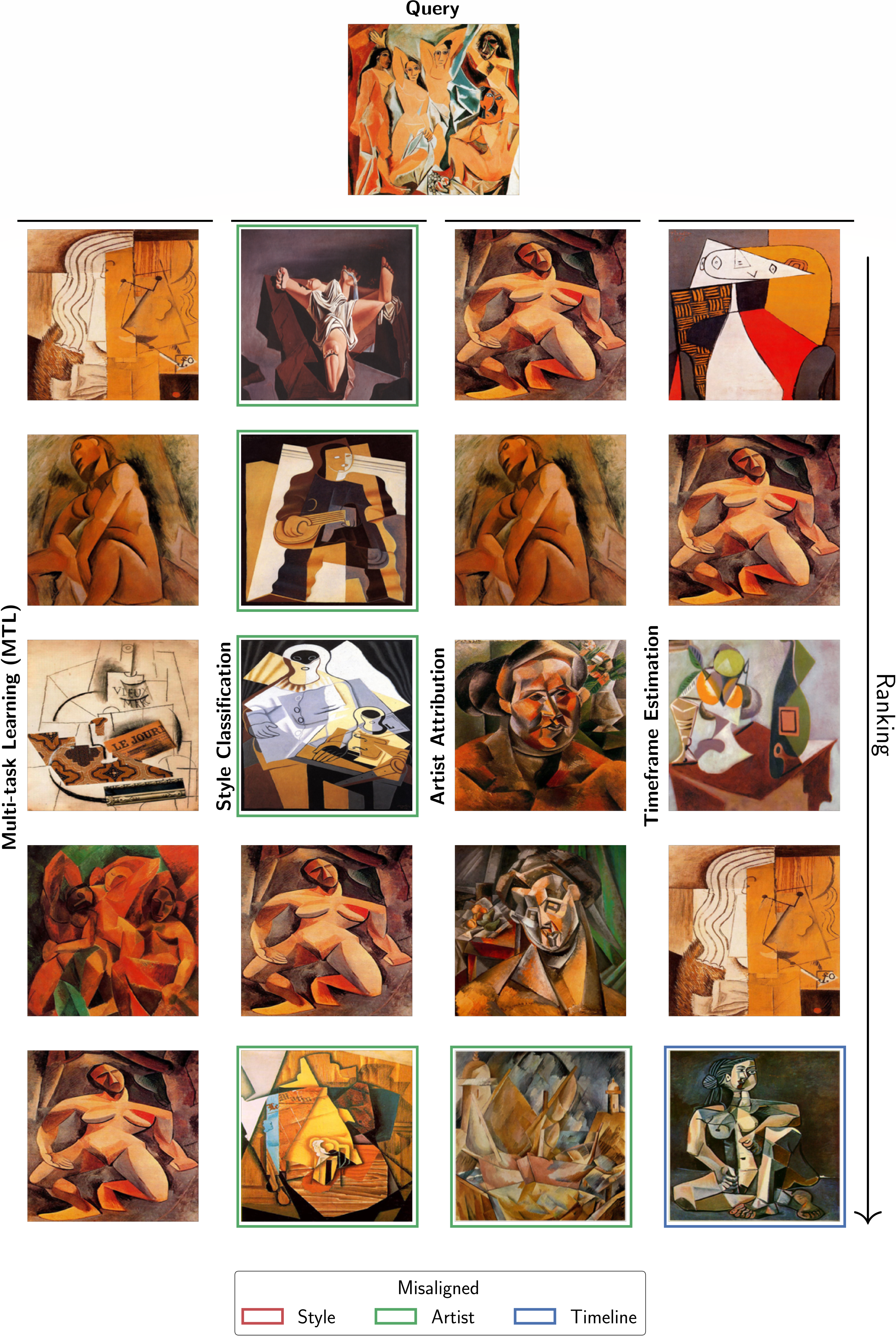}
\caption{Qualitative analysis of single and multi-task ArtSAGENet learned representations for painting retrieval.}
\label{fig:query_sagenet}
\end{figure}

Figure \ref{fig:query_sagenet} depicts the top-5 nearest neighbors of Pablo Picasso's proto-cubist \textit{Les Demoiselles d'Avignon} (1907) painting for single-task and multi-task ArtSAGENet. It is worth pointing out that all the retrieved artworks have the same relevant attribute, except for the artist attribution's fifth nearest neighbor, which is the \textit{Little Harbor in Normandy} (1909), one of the first examples of Georges Braque's early cubist style, and the fifth nearest neighbor for timeframe estimator, which is Pablo Picasso's \textit{Seated Woman} (1953). Finally, yet another interesting observation is about the top-5 paintings returned from single-task ArtSAGENet style classifier. We can see that the first nearest neighbor is Pablo Picasso's fellow Spanish countryman Salvador Dalí's \textit{Figura damunt les roques} (1926) cubist painting, while the second, third, and fifth nearest neighbors belong to another great Spanish artist, Juan Gris, who was closely connected with the emergence of the Cubism art movement, and Pablo Picasso's friend and neighbor in Montmartre, Paris.

\section{Conclusion}
\label{conclusion}

In this work, we proposed ArtSAGENet, a multimodal approach that learns knowledge enhanced visual representations of fine art. Experimental results illustrate that our proposed method leverages both visual and semantic-based information achieving state-of-the-art performance in several fine art categorization tasks. Qualitative analysis of the learned representations shows that ArtSAGENet encodes interesting semantic-level properties, which suggest, for instance, that Pablo Picasso's works in his Cubist period have a particularly close relation with those of Juan Gris. As this paper has demonstrated, integrating visual and semantic content provides a solid foundation for further research exploring such relationships in visual arts, and especially the dynamic processes of influence and innovation.

\balance

\bibliographystyle{ACM-Reference-Format}
\bibliography{references}

\clearpage

\appendix

\section*{Supplementary Material}\label{sec:supplementary_material}

In this supplementary material, we provide additional information for the dataset collection(Section~\ref{sec:dataset_collection}), ablation studies for the creation year estimation task(Section~\ref{sec:creation_year_estimation}), comparative results for from-scratch vs. fine-tuned trained ResNet-152 model(Section~\ref{sec:training_resnet_152}), sensitivity analysis for GraphSAGE model(Section~\ref{sec:sensitivity_analysis}), and results using alternate merging operators for our ArtSAGENet model(Section~\ref{sec:artsagenet_merging_operations}).

\section{Dataset Collection}
\label{sec:dataset_collection}

Here, we provide the necessary details for collecting the WikiArt dataset \cite{www.wikiart.org} and its variants. Since labeling artworks' stylistic categories has been shown to be quite demanding \cite{elgammal_art_history_machine}, we use the fine-grained obtained annotations from the WikiArt collection as in \cite{celtinic_fine_tuning_cnn, elgammal_art_history_machine, tan_artgan}. Table~\ref{tab:stylistic_categories} reports the number of paintings per stylistic category in the WikiArt\textsuperscript{Full} dataset. Furthermore, we gathered information regarding the artworks' respective artists and creation dates.  To alleviate the issue of missing annotations for creation dates, we created a \textit{timeframe} property, which is basically an attribute of half-centuries, e.g., from 1900 to 1950. We used the artists' active years, as provided from WikiArt, to automatically annotate artworks with missing dates to a timeframe. Finally, we map each artwork with a known creation date to its respective timeframe, e.g., for a painting that was created in 1907 the respective timeframe would be 1900-1950. For the tag prediction task, we gathered the tags associated with the paintings photographic reproductions on WikiArt collection. We make use of 54,919 paintings' images that are associate with at least one tag that, in turn, appears at least in 1,000 unique images. The latter results in a total of 54 unique tags, ranging from face parts, e.g., forehead and lips, nature related tags, e.g., flowers and animals, urban objects, e.g., vehicles and boats, to themes, e.g., children portraits and famous people.

\begin{figure}[h]
    \centering
    \includegraphics[width=\columnwidth]{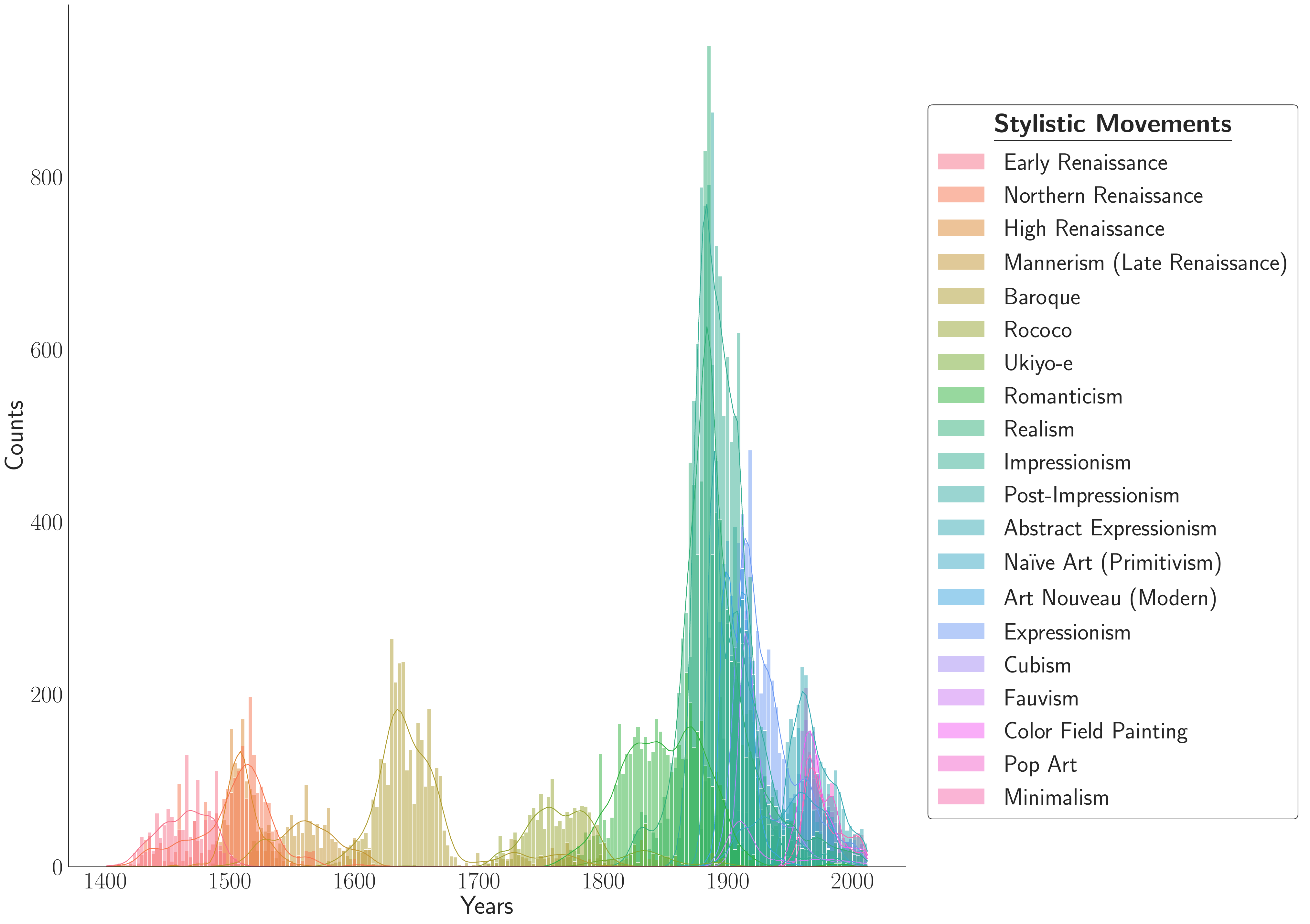}
    \caption{Stylistic Movements distribution over time on WikiArt dataset.}
    \label{fig:full_dataset_kde}
\end{figure}

Although the timeframe attribute as defined in the aforementioned paradigm can be an interesting property of an artwork, our goal is to build machines that can efficiently predict the exact creation year. As it can be clearly seen from Figure~\ref{fig:full_dataset_kde}, in the WikiArt 76k collection, there are approximately 50k artworks with a known creation date spanning a time period of 587 years. However, we experienced that annotations of artworks creation dates from early artistic movements, e.g., the Renaissance related movements, are inexact, with most cases being annotated with an estimated period rather a single year. In order to facilitate a more fine-grained evaluation of our approach and the baselines on creation date estimation task, we propose a subset of the WikiArt 76k collection that consists of all paintings present that are attributed to any given year from 1850. We call this subset, that consists of 45,869 artworks with a known creation date, WikiArt\textsuperscript{Modern}. Figure~\ref{fig:full_dataset_kde} shows the distribution of stylistic movements over time. It can be clearly seen that during the modern art period, there is a significant increase in emerging stylistic movements.

\begin{table}[h]
    \centering
    \caption{Stylistic categories in WikiArt\textsuperscript{Full}.}
    \label{tab:stylistic_categories}
    \begin{tabular*}{\columnwidth}{l @{\extracolsep{\fill}} S[table-parse-only,table-text-alignment=left]}
    \toprule
       Style  & {Number of Paintings} \\\midrule
        Early Renaissance & 1391\\
        Northern Renaissance & 2552\\
        High Renaissance & 1343\\
        Mannerism (Late Renaissance) & 1279\\
        Baroque & 4235\\
        Rococo & 2087\\
        Ukiyo-E & 1167\\
        Romanticism & 6957\\
        Realism & 11403\\
        Impressionism & 13019\\
        Post-Impressionism & 6811\\
        Abstract Expressionism & 2855\\
        Naive Art (Primitivism) & 2329\\
        Art Nouveau (Modern) & 4325\\
        Expressionism & 6563\\
        Cubism & 2528\\
        Fauvism & 811\\
        Color Field Painting & 1546\\
        Pop Art & 1449\\
        Minimalism & 1271\\\midrule
        Total Number of Paintings & 75921\\\bottomrule
    \end{tabular*}
\end{table}

Yet another issue that one may experience when utilizing the WikiArt dataset collection is that the representation of the artists within the dataset follows a Zipfian-type long tail distribution. \citet{tan_artgan, celtinic_dl_art} among others have previously used a subset of the WikiArt 76k collections to attribute artists to artworks. In order to be consistent with this line of work, and, since a large number of artworks within the WikiArt 76k dataset is attributed to only a few artists, we evaluate the performance of our models on a subset that, we call WikiArt \textsuperscript{Artists}, and consists of 17,785 available paintings of the 23 most representative artists within the WikiArt 76k collection. Table~\ref{tab:artist_distribution} reports the number of paintings per artist in the WikiArt\textsuperscript{Artists} subset.

\begin{table}[h]
    \centering
    \caption{Artists distribution in WikiArt\textsuperscript{Artists}.}
    \label{tab:artist_distribution}
    \begin{tabular*}{\columnwidth}{l @{\extracolsep{\fill}} S[table-parse-only,table-text-alignment=left]}
    \toprule
       Artist  & {Number of Paintings} \\\midrule
        Vincent van Gogh & 1890 \\
        Pierre-Auguste Renoir & 1400 \\
        Claude Monet & 1334 \\
        Pyotr Konchalovsky & 919 \\
        Camille Pissarro & 887 \\
        Albrecht D{\"u}rer & 828 \\
        John Singer Sargent & 784 \\
        Rembrandt & 777 \\
        Marc Chagall & 765 \\
        Gustave Doré & 753 \\
        Pablo Picasso & 745 \\
        Nicholas Roerich & 650 \\
        Boris Kustodiev & 633 \\
        Edgar Degas & 611 \\
        Paul Cézanne & 579 \\
        Ivan Aivazovsky & 577 \\
        Eugène Boudin & 555 \\
        Childe Hassam & 550 \\
        Ilya Repin & 539 \\
        Ivan Shishkin & 520 \\
        Raphael Kirchner & 516 \\
        Martiros Saryan & 510 \\
        Salvador Dalí & 463 \\\midrule
        Total Number of Paintings & 17785\\\bottomrule
    \end{tabular*}
\end{table}

\section{Creation Year Estimation}
\label{sec:creation_year_estimation}

\begin{figure}[h]
    \centering
    \includegraphics[width=\columnwidth]{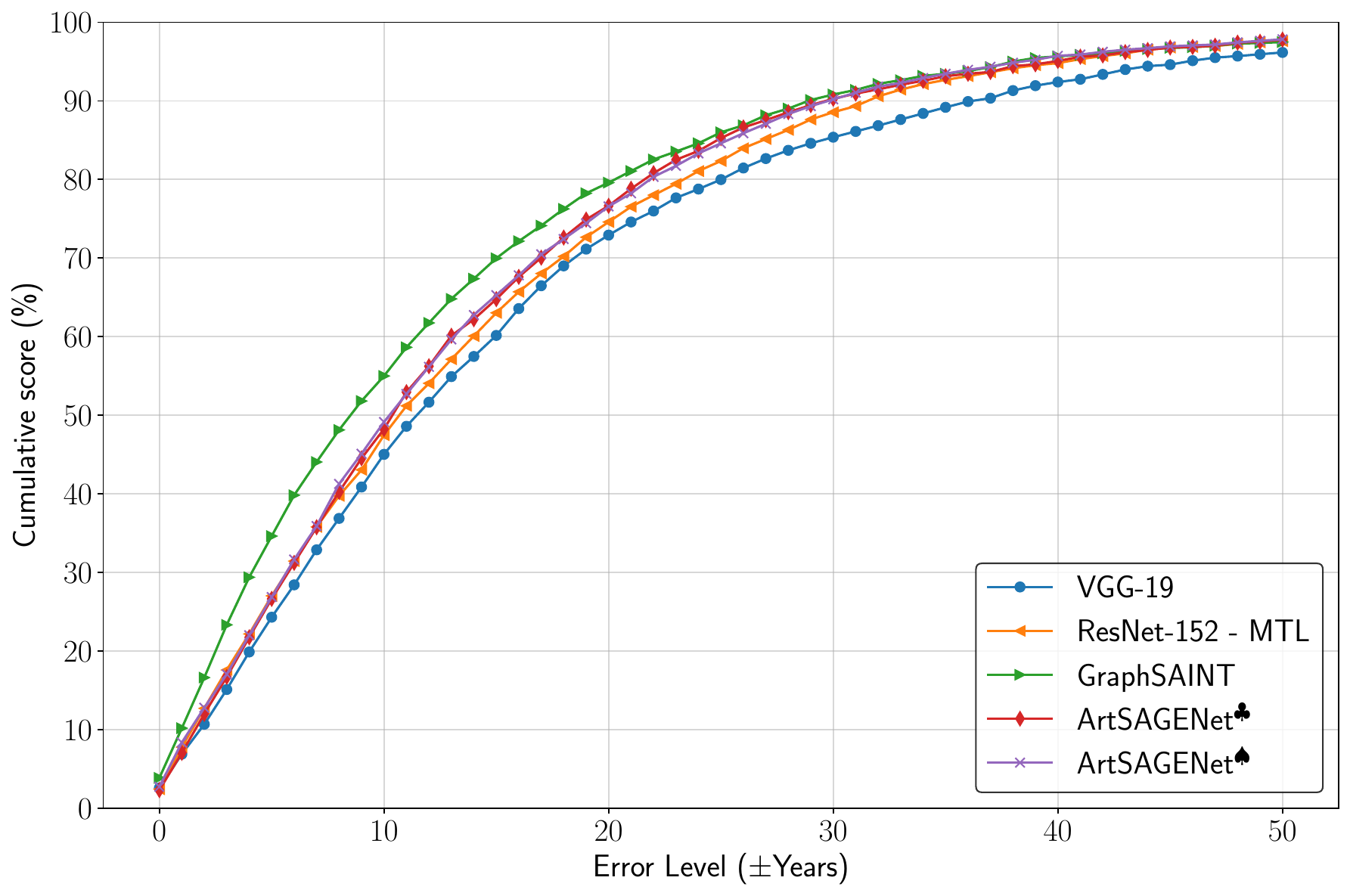}
    \caption{Cumulative score for the Creation Year Estimation task on the WikiArt\textsuperscript{Modern} dataset variant.}
    \label{fig:cumulative_interval_accuracy_regression}
\end{figure}

We follow the same practice as in human age estimation task~\cite{age_estimation_bridge_net, age_estimation_mean_variance} and evaluate the models performance on the creation year estimation task using the cumulative score~\cite{age_estimation_cumulative_score}. We compute the cumulative score as follows:

\begin{align} \label{al:cumulative_score_}
    CS(\theta) &= \frac{N_\theta}{N}\times100,
\end{align}

where $N$ is the total number of paintings in the test set and $N_\theta$ denotes the number of paintings whose absolute error is less than $\theta$ years.

Figure \ref{fig:cumulative_interval_accuracy_regression} depicts the performance of the five best models in terms of cumulative score. All models achieve an impressive performance in estimating the creation year within a $\pm20$ years period, while GraphSAINT can almost predict the artworks creation year with approximately 60\% accuracy within a $\pm10$ years period.

\begin{figure}[h]
\centering
\includegraphics[width=\columnwidth]{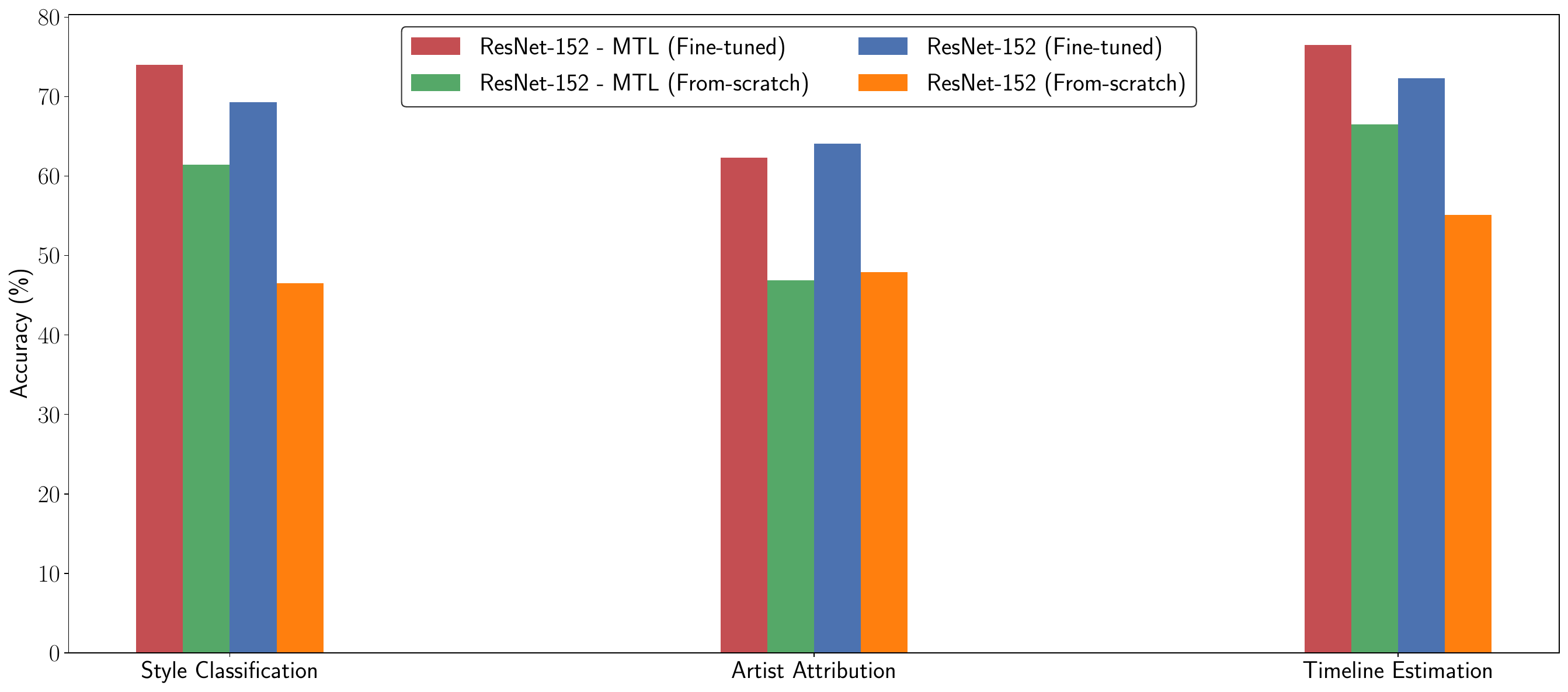}
\caption{Validation accuracy for ResNet-152 trained from-scratch or fine-tuned in the WikiArt\textsuperscript{Full} variant. MTL denotes Multitask Learning. Higher is better.}
\label{fig:resnet_152_from_scratch_vs_fine_tune}
\end{figure}

\begin{table*}[h]
\centering
\caption{Validation accuracy of ArtSAGENet$\spadesuit$ given four different merging operators. WikiArt\textsuperscript{Modern} - Date$\ddagger$ reports the cumulative score as in Eq. \eqref{al:cumulative_score} with $\theta=5$ years. $\spadesuit$ means using features extracted from a ResNet-34 model pre-trained on ImageNet as node feature vectors. Higher is better (best results in \textbf{bold}).}
\label{tab:merging_operations}
\begin{tabular*}{\linewidth}{l @{\extracolsep{\fill}} ccccccccc}
\toprule
                & \multicolumn{3}{c}{WikiArt\textsuperscript{Full}}                                                     & \multicolumn{3}{c}{WikiArt\textsuperscript{Modern}}                                                       & \multicolumn{3}{c}{WikiArt\textsuperscript{Artists}}                                                          \\ \cmidrule( r){2-4}
\cmidrule(lr){5-7}
\cmidrule(l ){8-10}
Model                    & Style & Artist & Timeframe & Style & Artist & Date$\ddagger$ & Style & Artist & Timeframe \\ \midrule
Add   &    77.0 & 74.9 & 78.7           &      76.1 & 75.4 & 19.8         &    88.8 & 97.4 & 89.1                     \\
Multiply  &    \textbf{78.9} & 74.3 & \textbf{81.1}        &   \textbf{77.4} & 75.1 & 19.4          &  \textbf{90.2} & 96.9 & \textbf{90.2}                          \\
Average      &      76.6 & 73.4 & 79.2            &      76.1 & 74.0 & 21.3     &       88.6 & 97.6 & 89.2                    \\
Concatenate     &             76.9 & \textbf{75.7} & 79.3                &     75.7 & \textbf{77.3} & \textbf{24.6}        & 88.6 & \textbf{98.0} & 88.0  \\\bottomrule
\end{tabular*}
\end{table*}

\begin{figure*}[hbt!]
\centering
\includegraphics[width=\textwidth]{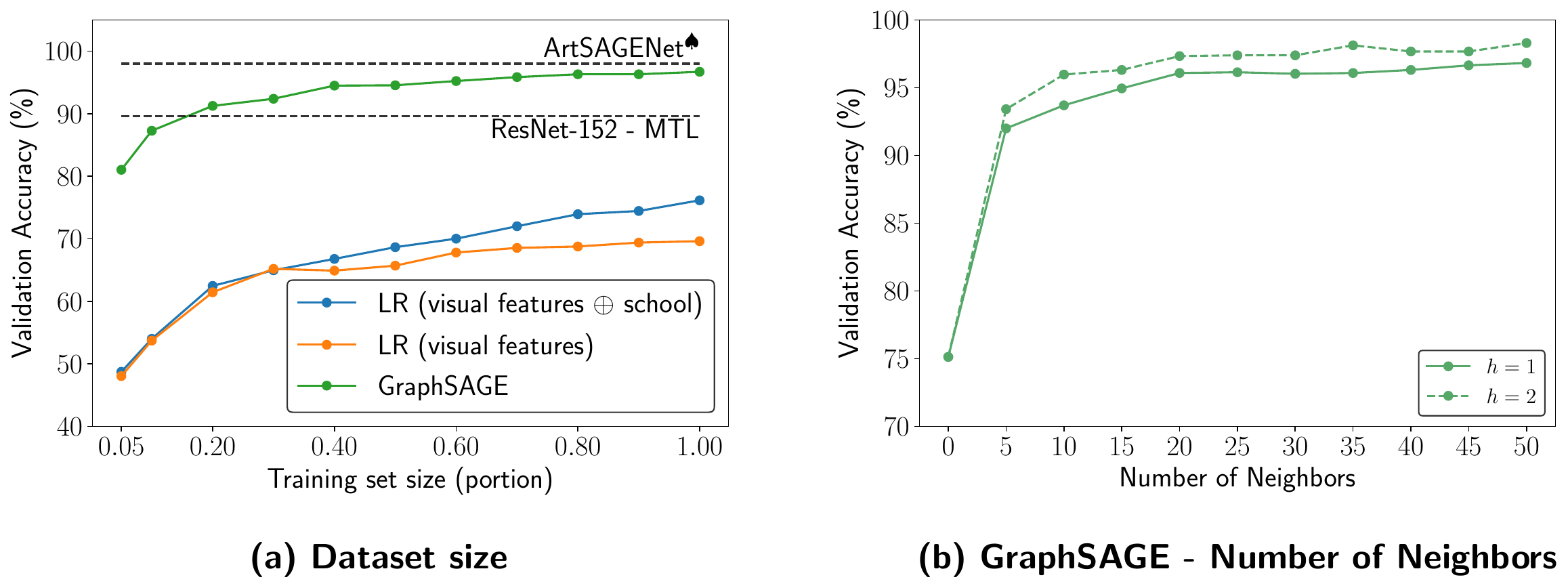}
\caption{GraphSAGE performance evaluation on WikiArt\textsuperscript{Artists} for the Artist Attribution task. (a) illustrates the validation accuracy of GraphSAGE with varying training set size compared to a simple Logistic Regression classifier. (b) illustrates the validation accuracy of GraphSAGE given the number of the neighbors and hops $h$.}
\label{fig:graphsage_performance}
\end{figure*}

\section{from-scratch vs. fine-tuned ResNet-152}
\label{sec:training_resnet_152}

For the traditional CNN baselines, we conducted experiments with both training from-scratch and fine-tuning pre-trained on ImageNet\cite{imagenet} models. Similar to \cite{elgammal_art_history_machine}, we found that the fine-tuned models achieved higher predictive performance in terms of accuracy compared to the from-scratch trained counterparts. Figure \ref{fig:resnet_152_from_scratch_vs_fine_tune} shows the classification accuracy for from-scratch and fine-tuned pre-trained on ImageNet ResNet-152 model on WikiArt\textsuperscript{Full} dataset. It can be easily observed that fine-tuning consistently outperforms training from-scratch for both single-task and multi-task settings.

\section{Sensitivity Analysis}
\label{sec:sensitivity_analysis}

In order to further validate the outstanding performance of GraphSAGE on the WikiArt\textsuperscript{Artists} artist attribution task, we conducted two additional experiments based on the training set size and the number of neighbors and hops taken into account. Figure \ref{fig:graphsage_performance} (a) depicts the effect of the available training data samples for GraphSAGE performance on WikiArt\textsuperscript{Artists} task. It can be easily observed that GraphSAGE achieves a remarkable performance by using only 20\% of the available labeled data for training. In addition, Figure \ref{fig:graphsage_performance} (a) shows that a simple Logistic Regression classifier using either visual features or both visual features and the art school attributes consistently under-performs GraphSAGE.

Moreover, Figure \ref{fig:graphsage_performance} (b) shows the effect of the number of available hops $h$ and the number of neighbors considered for GraphSAGE on the WikiArt\textsuperscript{Artists} artist attribution task. It can be clearly seen that GraphSAGE can effectively discriminate the artists using only one hop for each artwork. Nevertheless, we found that sampling from the second hop, $h=2$, improves the performance compared to sampling only from the first hop, $h=1$, for the artist attribution task. Finally, increasing the number of neighbors considered, positively effect the discrimination power of GraphSAGE, albeit it seems that the predictive performance reaches a plateau after considering more than 20 neighbors.

\section{ArtSAGENet Merging Operations}
\label{sec:artsagenet_merging_operations}

For our ArtSAGENet architecture we considered four different multi-modal vector composition operators, namely, addition, multiplication, averaging and concatenation. Table~\ref{tab:merging_operations} reports the evaluation of the predictive performance for the four composition operators. It is clear that multiplication and concatenation are both the best performing composition functions. However, we have to highlight that we found that concatenation is usually the most stable and fast in terms of convergence.

\end{document}